%% file: emnlp2022.tex
\pdfoutput=1
\documentclass[11pt]{article}

\usepackage[]{EMNLP2022}
\usepackage{times}
\usepackage{latexsym}

\usepackage[T1]{fontenc}
\usepackage[utf8]{inputenc}

\usepackage{microtype}

\usepackage{inconsolata}

\usepackage{subcaption}

\newcommand{\dn}{\abr{C3}\xspace} % dataset name
\title{Tell Your Story: Task-Oriented Dialogs for Interactive Content Creation}

\include{tex/simmc_macros}

\include{tex/space_saver}
\include{tex/shane_macros}

\usepackage[misc]{ifsym} % Letter
\newcommand*\samethanks[1][\value{footnote}]{\footnotemark[#1]}

\author{
    Satwik Kottur \thanks{\hspace{5pt}Joint first authors},
    Seungwhan Moon \samethanks,
    Aram H. Markosyan,\\
    \textbf{Hardik Shah,
    Babak Damavandi,
    Alborz Geramifard} \\
    Meta Reality Labs \& Meta AI\\
    {\small \Letter}\hspace{3pt} \texttt{\{skottur,shanemoon,amarkos,hjshah,babakd,alborzg\}@fb.com} \\
}

\date{}

\begin{document}
\maketitle
\begin{abstract}
\sk{
    People capture photos and videos to relive and share memories of personal significance.
    Recently, media montages (stories) have become a popular mode of sharing these memories
    %with friends and family 
    due to their intuitive and powerful storytelling capabilities.
    However, creating such montages usually involves a lot of manual searches, clicks, and selections 
    that are time-consuming and cumbersome, adversely affecting user experiences.
    
    To alleviate this, we propose \textit{task-oriented dialogs for montage creation} as a novel
    interactive tool to seamlessly search, compile, and edit montages from a media collection.
    To the best of our knowledge, our work is the first to leverage multi-turn conversations for such a
    challenging application, extending the previous literature studying simple media retrieval tasks.
    We collect a new dataset \dn{} (\textbf{C}onversational \textbf{C}ontent \textbf{C}reation), comprising $10k$ dialogs conditioned on media montages simulated from
    a large media collection.

    We take a simulate-and-paraphrase approach to collect these dialogs to be both cost and time 
    efficient, while drawing from natural language distribution.
    Our analysis and benchmarking of state-of-the-art language models showcase the 
    multimodal challenges present in the dataset.
    Lastly, we present a real-world mobile demo application that shows the feasibility of the proposed work in real-world applications.
    Our code \& data will be made publicly available.
}
\end{abstract}

\section{Introduction}
\label{sec:introduction}
\vspace{-4pt}
\input{sections/introduction}

\vspace{-15pt}
\section{Related Work}
\label{sec:related_work}
\vspace{-4pt}

\input{sections/related_work}

%\vspace{-2pt}
\section{The \dn Dataset}
\label{sec:dataset}
\vspace{-4pt}
\input{sections/dataset}

%\subsection{\dn Dataset Analysis}
%\label{sec:dataset_analysis}
%\input{sections/dataset_analysis}

\vspace{-20pt}
\section{Task Formulation}
\label{sec:task_formulation}
\vspace{-6pt}
\input{sections/task_formulation}

\vspace{-2pt}
\section{Modeling \& Empirical Analysis}
\label{sec:modeling}
\vspace{-4pt}
\input{sections/modeling}

% \section{Conclusions}
% \label{sec:conclusions}
\input{sections/conclusions}

\section{Limitations}
\label{sec:limitations}

\input{sections/limitations}

\bibliography{bibliography}
\bibliographystyle{acl_natbib}

\vfill % enforce filling the half filled column
\eject
\columnbreak

\clearpage % enforce an empty column

\appendix
\input{sections/supplementary}

\end{document}

%% file: tex/simmc_macros.tex
\newcommand{\reffig}[1]{Fig.~\ref{#1}}
\newcommand{\refsec}[1]{Sec.~\ref{#1}}
\newcommand{\reftab}[1]{Tab.~\ref{#1}}

\newcommand{\reportval}[2]{$#1$\footnotesize $\pm #2$}

\usepackage{booktabs,arydshln}
\makeatletter
\def\adl@drawiv#1#2#3{%
        \hskip.5\tabcolsep
        \xleaders#3{#2.5\@tempdimb #1{1}#2.5\@tempdimb}%
                #2\z@ plus1fil minus1fil\relax
        \hskip.5\tabcolsep}
\newcommand{\cdashlinelr}[1]{%
  \noalign{\vskip\aboverulesep
           \global\let\@dashdrawstore\adl@draw
           \global\let\adl@draw\adl@drawiv}
  \cdashline{#1}
  \noalign{\global\let\adl@draw\@dashdrawstore
           \vskip\belowrulesep}}
\makeatother

%% file: tex/space_saver.tex
\belowdisplayskip \abovedisplayskip

\newlength{\sectionReduceTop}
\newlength{\sectionReduceBot}
\newlength{\subsectionReduceTop}
\newlength{\subsectionReduceBot}
\newlength{\abstractReduceTop}
\newlength{\abstractReduceBot}
\newlength{\captionReduceTop}
\newlength{\captionReduceBot}
\newlength{\subsubsectionReduceTop}
\newlength{\subsubsectionReduceBot}

\newlength{\eqnReduceTop}
\newlength{\eqnReduceBot}

\newlength{\horSkip}
\newlength{\verSkip}

\newlength{\figureHeight}

%% file: tex/shane_macros.tex
\usepackage{graphicx} 
\usepackage{subcaption}
\usepackage{multicol}
\usepackage{booktabs} 
\usepackage{algorithm} 
\usepackage{algorithmic} 
\usepackage{color}
\usepackage{amsmath}
\usepackage{amssymb}
\usepackage{amsfonts}
\usepackage{multirow, varwidth}
\usepackage{stfloats} 
\usepackage{verbatim}
\makeatletter
\newif\if@restonecol
\makeatother
\usepackage{xr}
\usepackage[amsmath,thmmarks]{ntheorem}

\usepackage{xspace}
\makeatletter
\DeclareRobustCommand\onedot{\futurelet\@let@token\@onedot}
\def\@onedot{\ifx\@let@token.\else.\null\fi\xspace}

\def\eg{\emph{e.g}\onedot} 
\def\ie{\emph{i.e}\onedot}

\makeatother

\newcommand{\emptydraft}[1]{{\color{black}{}}}

\newcommand{\sk}[1]{{\color{black}{#1}}}
\usepackage{xcolor}
\usepackage{enumitem}
\usepackage{flushend}
\newcommand{\abr}[1]{\textsc{#1}}

%% file: sections/introduction.tex
\begin{figure}[t!]
    \centering
    \includegraphics[width=0.73\columnwidth]{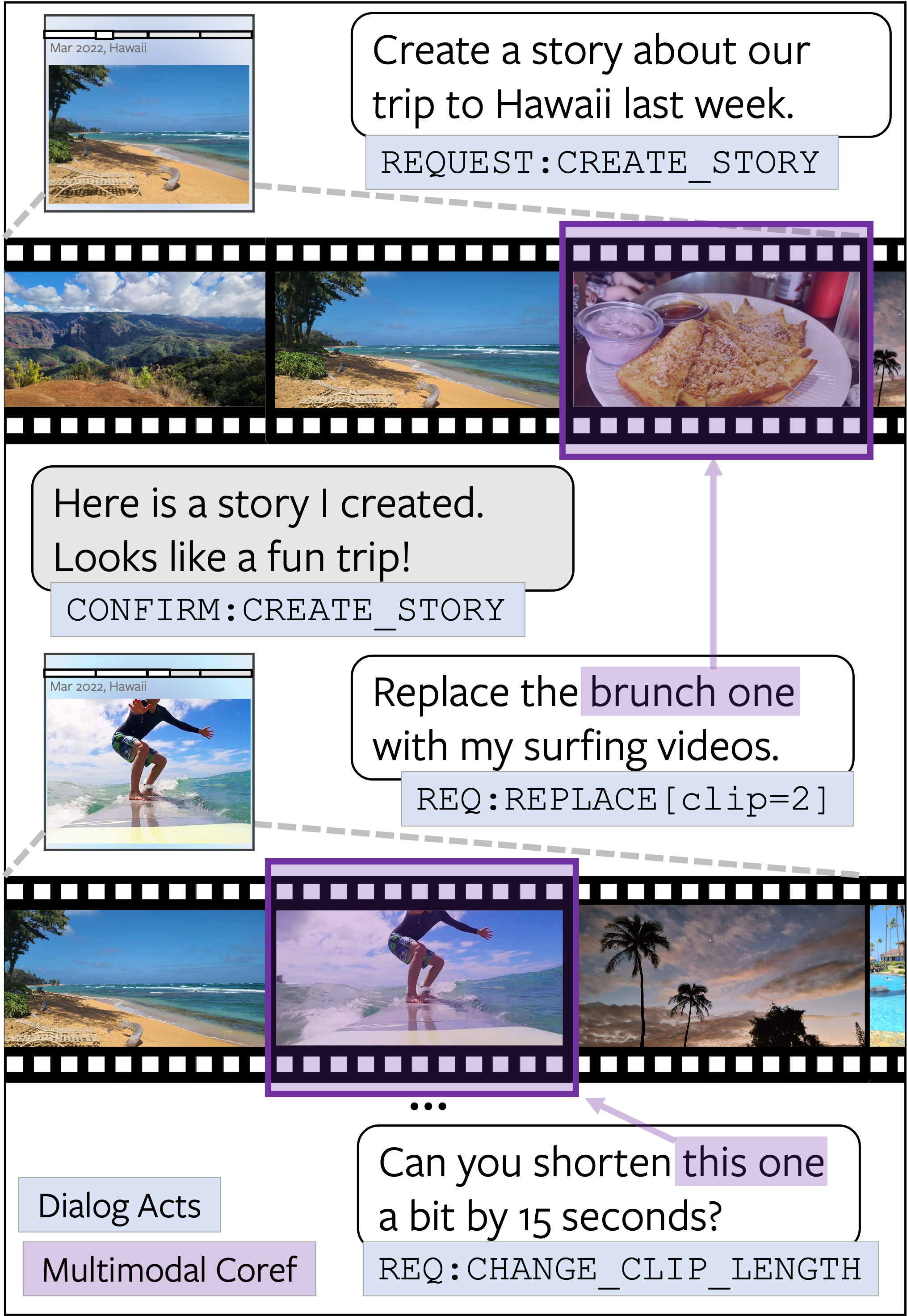}
    %\vspace*{\captionReduceTop}
    \vspace{-4pt}
    \caption{Illustration of \dn: \textbf{C}onversational \textbf{C}ontent \textbf{Creation}. Each dialog turn is fully annotated with dialog acts and multimodal coreference labels, accompanied with its corresponding story montage snapshot.}
    %\vspace*{\captionReduceBot}
    \vspace{-16pt}
    \label{fig:teaser}
\end{figure}

\sk{
    With the advent of smart cameras, smart glasses, and other media devices, the barrier to capturing 
    photos and videos has drastically been reduced.
    While this trend is desirable to relive and share memories, the sheer volume
    of such captured media makes it intractable to search and share relevant memories.
    As a result, media montages (stories) have emerged as an intuitive yet expressive way to creatively
    compile various memories and share with friends and family.
    In order to create such a montage, users have to search through their personal collections,
    make selections, and edit them manually, which are cumbersome and time-consuming tasks, 
    resulting in a bottleneck.
}

\sk{
     In this work, we propose a novel conversational tool to interactively create
     and edit montages from a personal media collection.
     While prior works study the use of dialog in retrieving media or items in a shopping catalog, 
     we extend it to capture richer interactions related to montage manipulations.
     To the best of our knowledge, our work is the first to consider task-oriented dialogs (TOD) for this 
     challenging application of interactive content creation.
}

\sk{
    Towards this goal, we collect \dn{}, a TOD dialog dataset aimed at providing an intuitive 
    conversational interface in which users can search through their media, create a video story with
    highlights, and edit clips hands-free, using natural language.
    \reffig{fig:teaser} illustrates an example dialog.
    Due to our simulate-and-paraphrase pipeline, our dataset comprises rich annotations both at
    turn- and dialog-level.
    These are helpful to: (a) tease out and study multimodal challenges (\eg, multimodal coreferences)
    that are present in \dn, and (b) benchmark meaningful progress towards a robust 
    TOD agent for this application.
    We perform preliminary empirical experimentation and train baselines to highlight the multimodal
    challenges in our \dn dataset.
    % Putting it all together, 
    Lastly, we build a mobile demo (\reffig{fig:demo}, App. \ref{appendix:demo}) to showcase
    the real-world applicability of our work.
    % help people frictionlessly create shareable narratives via a conversation. 
}

%% file: sections/related_work.tex
\noindent \textbf{Task-oriented Dialogs} (TOD), where the goal is to parse user queries and execute a pre-defined set of actions (\eg booking hotels), have been extensively studied. %due to the popularity of virtual assistants. %consumer-facing 
We formulate similar tasks as found in the conventional TOD datasets \cite{sgd-dst,multiwoz,multiwoz2.1} such as Dialog State Tracking (DST), to build on the literature. %Natural Language Understanding (NLU), 
Our work extends it to a novel multimodal application of video content creation and editing.

Recently, the methods that leverage large pre-trained LMs by casting DST as a causal inference problem \cite{soloist,simpletod,bert-dst-alexa} have shown successful.
%(\eg GPT-2 \cite{radford2019language}) 
We develop a baseline following this trend, but extend it a unique multimodal setting by including multimodal context as part of the grounding prompt. 

\noindent \textbf{Conversational Media Applications}: 
Recent work have addressed the dialog task for retrieving images (\eg from a personal collection or as part of shopping scenarios) \cite{dialog-image-retrieval,NEURIPS2018_a01a0380,10.1145/1646396.1646442,Vo_2019_CVPR,drilldown}, given multi-turn target queries. %, where the goal is typically to improve media retrieval performances given target queries.
Similarly, \citet{bursztyn2021gaud} considers an application to retrieve multiple images to create a montage.
While \dn{} does include search operations, our work extends this line of work by allowing for richer interactions and more complex post-edits on the retrieved videos, enhancing overall user experiences.

%Another similar applications include the work by \citet{lin2020multimodal}, which proposes a task for editing a single image (\eg brightness, saturation) via language commands.
As per similar applications, \citet{lin2020multimodal} proposes tasks for editing a single image (\eg brightness) via text commands, while
\citet{zhou2022interactive} study interactive image generation from text, using CLIP text-image embeddings \cite{clip} and a generative model \cite{stylegan}.
Unlike the previous work that handle editing operations within a single image, our work addresses conversational editing of %and compiling of
multiple videos into storytelling montages, a popular form of media sharing.

%% file: sections/dataset.tex
\subsection{Multimodal Dialog Self-Play}

% -- \refsec{sec:dataset:dialog_simulator} and \refsec{sec:dataset:paraphrase}
We adopt a two-phase pipeline (Simulate and Paraphrase \cite{shah2018building, simmc2}), extending it to a unique multimodal setting where multiple images as part of the user interface (UI) are given as grounding visual contexts.
%This approach is known to significantly 
The proposed approach reduces the data collection and annotation overheads (time and cost) for building a dialog dataset (\textit{vs.} collecting human$\leftrightarrow$human dialogs and collecting Dialog/NLU annotations on top), as it requires little to no domain knowledge.% on the annotator's part.

%We provide the in-depth analysis of the resulting \dn{} dataset in \refsec{sec:dataset_analysis}.

%\label{sec:dataset:dialog_simulator}
\noindent \textbf{Phase 1. Multimodal Dialog Simulator.}
We first generate synthetic dialog flows using a dialog simulator that conditions on an evolving “story” and its corresponding set of clips, produced by a story generator.
The story generator outputs a diverse set of clips (as schematic representation) according to user requests, which serves as grounding multimodal context for the conversations.
This is done by extracting a plausible set of meta information (time, locations and activities, etc.) from an existing memory graph, simulated and generated using the object and activity annotations from the ImageCOCO dataset \cite{mscoco}.

The dialog simulator then takes this story representation including the meta information (activities, locations, attributes, etc.) and the UI state (\eg sequential ordering of media, viewer status) updated at each turn, to create a realistic dialog flow between a user and an assistant, using a probabilistic agenda-based approach.
The simulated dialog flows comprise NLU intents (\eg \texttt{REQUEST:ADD\_CLIPS}), slots (\eg activities, objects), and clip references. %, following the ontology widely used in other multimodal ToD datasets \cite{simmc2}. % Natural Language Understanding (NLU)
Specifically, we capture various video editing queries that are identified as a prioritized list of common actions required for media editing and sharing (\eg \texttt{CREATE}, \texttt{REMOVE}, \texttt{REPLACE}, \texttt{REORDER}, \texttt{REFINE}, \texttt{MODIFY\_DURATION}).

%\subsubsection{Manual Paraphrase}
%\label{sec:dataset:paraphrase}
\noindent \textbf{Phase 2. Manual Paraphrase.}
Once the dialog flows are simulated, we paraphrase each templated user turn via manual annotations. % with the help of human annotators
This step allows us to collect utterances from the natural language distribution, making the dataset robust to the user-query variability in real-world applications.

%For the paraphrase task, w
We build an annotation tool that displays NLU labels and templated utterances, along with the schematic representation of stories with media clips, updated at each turn.
Annotators are then instructed to paraphrase each turn without losing key multimodal information such as relative clip placements \& meta data, objects and attributes.

%\todo{phrase it differently - that we also include assistant responses, as opposed to we only collected 10 percent.}
While assistant turns tend to be linguistically less diverse (\eg, informing successful executions: \textit{`Done'}, \textit{`Edited'}) and thus are less of our focus from an application standpoint, we also collect assistant responses for a 1$k$ dialog subset. 
The collected utterances allow for the study of contextualized assistant response generation, to accompany the modified stories reflected in the UI.
% original sentences
%Due to the nature of the task, assistant turns to inform successful execution of a contextual user instruction are linguistically less diverse (\eg, \textit{`Done'}, \textit{`Edited'}). 
%Further, these add little information from an application standpoint as the user can see the modified story based on their request.
%Thus, we collect assistant responses only for a subset of $1k$ dialogs ($10\%$).

\subsection{Dataset Analysis}
\label{sec:dataset:analysis}
\input{sections/dataset_analysis}

%% file: sections/dataset_analysis.tex
\sk{
Our \dn{} dataset has a total of $10k$ dialogs with $136k$ utterances.
%, which are grounded in $10k$ stories.
Dataset statistics are given in \reftab{tab:dataset_statistics}.
A dataset example is provided in \reffig{fig:example} (Appendix \ref{appendix:dataset_example}).
}

\begin{figure*}[t]
    \centering
    \begin{subfigure}[b]{0.25\textwidth}
        \includegraphics[width=\textwidth]{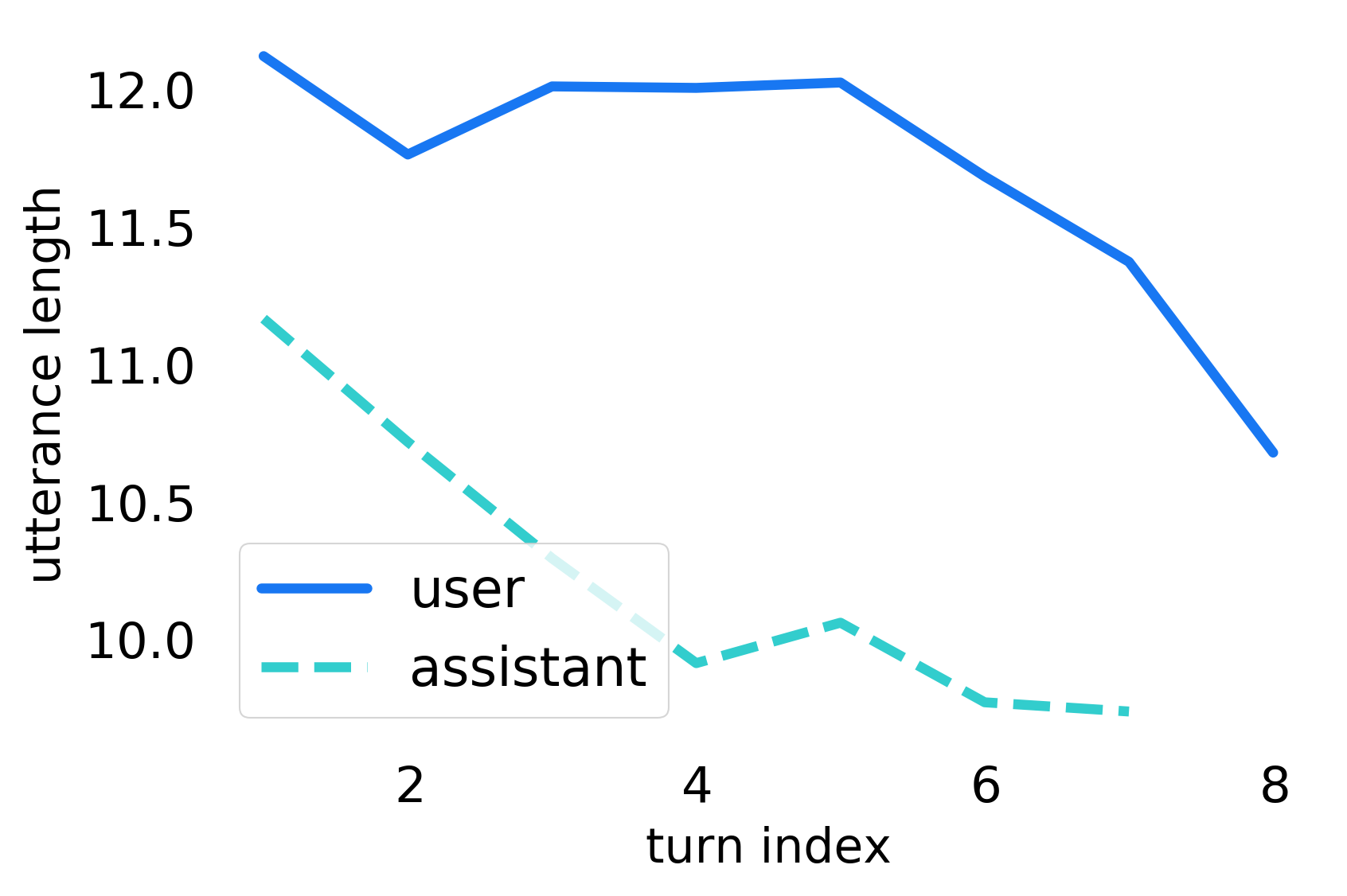}
        \caption{}
        \label{fig:utterance_len_distr}
    \end{subfigure}
    ~
    \begin{subfigure}[b]{0.3\textwidth}
        \includegraphics[width=\textwidth]{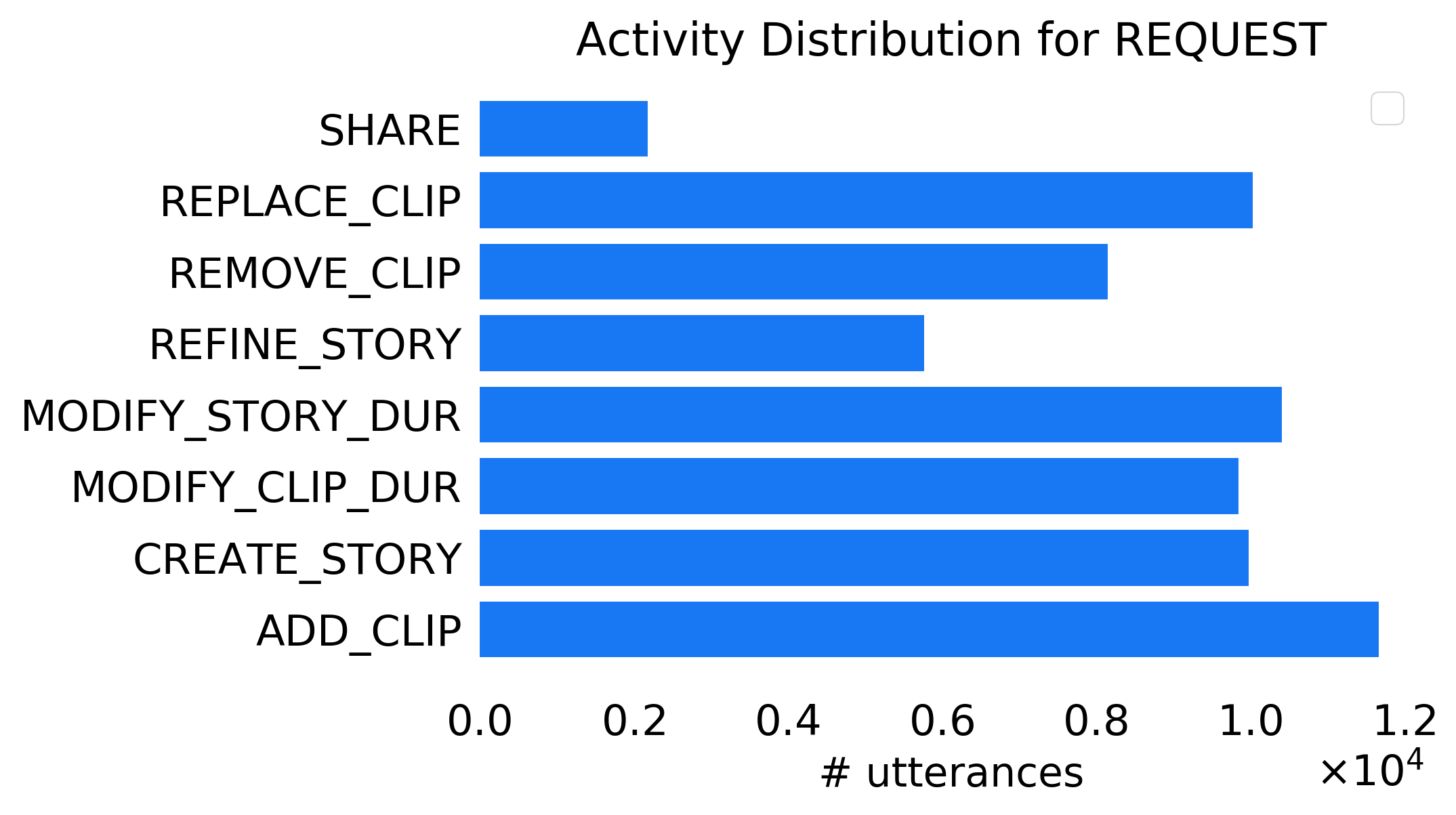}
        \caption{}
        \label{fig:api_distr}
    \end{subfigure}
    ~
    \begin{subfigure}[b]{0.40\textwidth}
        \includegraphics[width=\textwidth]{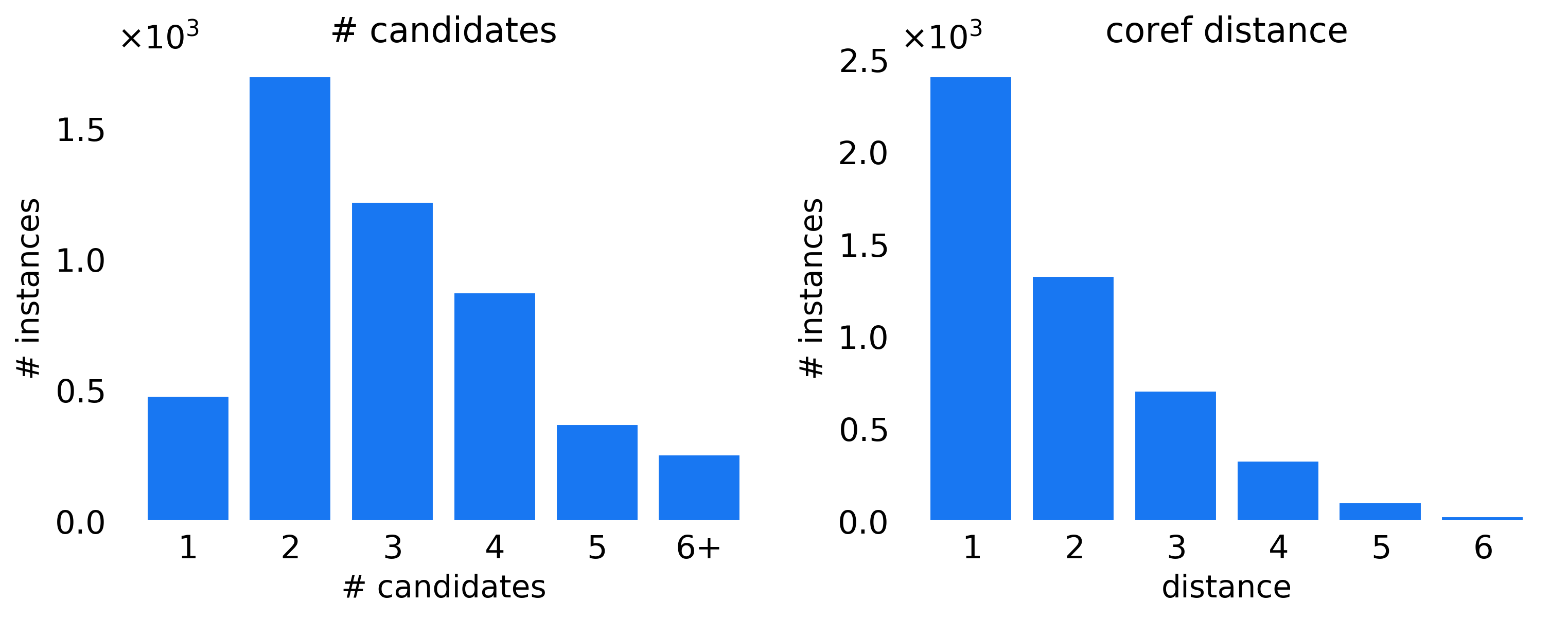}
        \caption{}
        \label{fig:coref_distr}
    \end{subfigure}
    \vspace*{\captionReduceTop}
    \caption{Distribution of 
    (a) utterance lengths with dialog turns,
    (b) activity distribution for \texttt{REQUEST} user act (dominant),
    (c) number of clip candidates per turn (L) and coreference distance (R) between clip mentions.
    }
    \vspace*{\captionReduceBot}
\end{figure*}

\begin{figure*}
    \centering
    \includegraphics[width=0.82\textwidth]{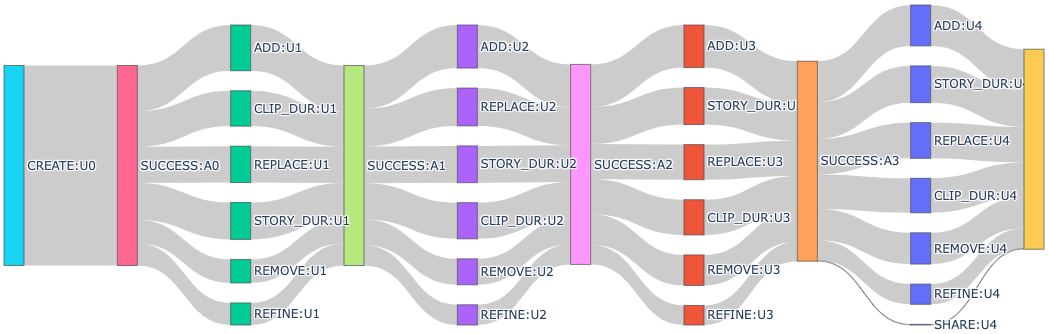}
    \vspace*{\captionReduceTop}
    \caption{
        Transition of dialogs acts in \dn for the first $4$ turns, for dialog flows generated by our
        multimodal dialog simulator.
        Each block is labelled \texttt{ACTIVITY:[A|U][turn]} to denote activity, 
        user or assistant turn, and turn number, respectively.
        \texttt{ACT} for user (\texttt{REQUEST}) and assistant (\texttt{INFORM}) are not shown for brevity.
        See text for more details.
    }
    \vspace*{\captionReduceBot}
    \label{fig:intent_transitions}
\end{figure*}

\input{tables/tab_dataset_statistics}

%\vspace*{-12pt}
\noindent
\textbf{Analyzing Dialogs.}
\sk{
    The user and assistant turns in dialogs from \dn{} are about $11.8$ and $10.3$ words long respectively,
    with their distributions shown in \reffig{fig:utterance_len_distr}.
    %As expected, 
    User utterances tend to be longer on an average as they are %  than assistant responses
    instructive and contain finer details to manipulate the story.
}

\noindent
\textbf{Analyzing Dialog Annotations.}
\sk{
    Dialogs in \dn{} are accompanied with full turn- and dialog-level annotations, thanks to the
    simulate-and-paraphrase approach.
    We follow the conventional hierarchical ontology \cite{simmc2} of dialog \texttt{ACT} and \texttt{ACTIVITY} to 
    annotate both user and assistant intents.
    %Recall that our motivation is to provide a conversational tool where 
    In our setup, users can \textit{request} selections or 
    edits to create a montage, while the assistant is expected to execute them and \textit{inform} its
    results.
    Thus, the user and assistant dialog acts naturally resort to \texttt{REQUEST} and \texttt{INFORM}
    in our ontology.
    \reffig{fig:api_distr} shows the distribution of 8 user activities.
    %, showing a reasonable distribution over $8$ values.
    %\texttt{SHARE} understandably occurs less frequently as it is typically the culminating dialog activity in each interaction.
}

\sk{
    Each turn is grounded on an (evolving) story, which contains an average of $4.3$ clips.
    This leads to interesting multimodal coreferences as there are about $2.9$ clip 
    candidates to pick from for every clip mention in the dialog.
    Further, the average coreference distance between the mentions is $3.7$, going beyond the trivial
    case of $1$, \ie, clip mentioned in the previous turn.
    \reffig{fig:coref_distr} highlights the distribution of clip candidates and distance between mentions.
}

\noindent
\textbf{Analyzing Dialog Flows.}
\sk{
%Finally, 
We visualize the dialog flows (first 4 dialog turns) in
%generated by the multimodal dialog simulator in 
\reffig{fig:intent_transitions}.
Each block is an intent at a particular \texttt{[turn]} labelled as \texttt{ACTIVITY:[A|U][turn]}, 
where \texttt{[A|U]} indicates either an \texttt{A}ssistant or \texttt{U}ser turn.
The gray bands denote the transitions and their width is proportional to the frequency of the transition.
The almost uniform branch-off indicates a desirable presence of diversity and thus a lack of intent bias in the dialog.
}

%% file: tables/tab_dataset_statistics.tex
\begin{table}[t]
    \begin{center}
        \scalebox{0.78}{
        \begin{tabular}{lccc}
        \toprule[\heavyrulewidth]
        Total \# dialogs & $10k$ \\
        Total \# utterances & $136k$ \\
        Total \# stories & $10k$ \\
        Avg \# words (user turns) & $11.8 \pm 4.4$ \\
        Avg \# words (assistant turns)$^\dagger$ & $10.3 \pm 4.1$ \\
        Avg \# utterances / dialog 	& $13.5$ \\
        Avg \# clips mentioned / dialog &	$3.6$ \\
        Avg \# clips per story & $4.3 \pm 2.5$ \\
        \bottomrule[\heavyrulewidth]
        \end{tabular}
        }
    \end{center}
    %\vspace{-10pt} 
     \vspace*{\captionReduceTop}
    \caption{
        \textbf{\dn{} Dataset Statistics.} 
        $^\dagger$assistant turns are collected for a 1$k$ dialog subset (12$k$ utterances).
    }
    \vspace*{\captionReduceBot}
    \vspace{-6pt}   
    \label{tab:dataset_statistics}
\end{table}

%% file: sections/task_formulation.tex
\sk{
    We leverage \dn{} to study dialog systems that help users create and edit
    montages through a multi-turn dialog.
    More concretely, we propose 3 main tasks and respective evaluation metrics in this regard:
}

\noindent
\textbf{Task 1: API Slot Prediction.}
\sk{
%In our work, w
We assume a 1-to-1 mapping between user intent and the relevant API to execute a user request.
API Slot Prediction thus involves predicting slots (e.g., \textit{participants}, \textit{time}) that are
passed as arguments to the corresponding API, given dialog history, multimodal context of stories,
and current user utterance (metric: F1).
For example, \textit{`U: Create a story of all skiing trips in 2018'} maps to
\texttt{[activity=skiing,}\texttt{time=2018]} as the appropriate API slots and values.
%We use slot F1 to measure and compare performances of models in this task.
We do not propose a separate API-type prediction task (\eg \texttt{api\_type=CREATE\_STORY}) as the baseline models perform with near perfect 
accuracy ($97\%$).
}

\noindent
\textbf{Task 2. Multimodal Coreference Resolution.}
\sk{
%As mentioned in \refsec{sec:dataset_analysis}, \dn{} contains coreference annotations that relate clip 
%mentions in the dialog utterance to the canonical clip objects in the story.
%It is imperative for conversational systems to be able to resolve these multimodal references without
It is imperative for conversational systems to be able to resolve multimodal coreferences without
fails as a wrongly targeted edit would require additional interactions to rectify, greatly reducing user experiences.
For instance, to process \textit{`Remove the \underline{sunset clip} and replace it with 
something \underline{similar to the second one}.'}, the system needs to resolve both underlined references
to the corresponding clip objects to perform the desired manipulations.
To test this capability in isolation, we propose Task 2, where the goal is to 
resolve any clip references in the current user utterance to the corresponding clip objects (metric: F1), taking into
account dialog history and story representations.
%Similar to \cite{simmc2}, we use F1 score to measure performance.
}

%\vspace{-2pt}
%\label{sec:dataset_analysis}
\noindent
\textbf{Task 3. Multimodal Dialog State Tracking (MM-DST).}
\sk{
%Putting it all together,
%We propose a multimodal dialog state tracking task where the system is evaluated on 
Lastly, we evaluate the system on its joint ability to:
(a) predict API calls along with its slot parameters,
and (b) resolve multimodal references (if any) in the given utterance,
taking into account dialog state carryovers (measured with accuracy).
%We use accuracy as a metric to compare different approaches in this task and benchmark progress.
}

%% file: sections/modeling.tex
\input{tables/tab_model_results}
We perform a preliminary empirical evaluation and train baselines for the tasks proposed in \refsec{sec:task_formulation}.
We leave detailed modeling as part of future work.

\noindent
\textbf{Dataset Splits.}
We split the $10k$ dialogs into \texttt{train} ($60\%$), \texttt{val} ($20\%$),
and \texttt{test} ($20\%$). All models are trained on \texttt{train} with \texttt{val} used to pick the
hyper-parameters, and results are reported on \texttt{test}.

\noindent 
\textbf{Baselines.}
Following the recent success of finetuning pretrained LMs on
TODs \cite{simpletod,soloist}, we adopt GPT-2 \cite{radford2019language} and extend these work by adding two different ways of representing multimodal contexts (story):
(a) visual embeddings (embed), where we extract object-centric visual features for constituent clips \cite{fasterrcnn} projected into the hidden size of GPT-2 via a linear layer, and
(b) stringified text (tokens), where the story information is represented as stringified tokens.
The models are trained to predict API calls, slot values, and clip mentions % (for coreference resolution)
given a sequential input of its dialog context and multimodal context as above, through a conditional LM loss.
More details are in Appendix \ref{appendix:baseline}. 

\noindent
\textbf{Results.}
From \reftab{tab:results},
it can be seen that the models achieve reasonably reliable performances for API prediction, while the coreference resolution task (exactly pinpointing which set of clips a user mentions) still remains a challenge.
This is due to the various types of coreferences that exist in \dn that make resolutions uniquely challenging (\eg adjectival: ``\textit{\underline{the sunset clip}}", ordinal: ``\textit{\underline{the second to the last one}.}, device context: ``\textit{\underline{the one I'm currently viewing}}", long-range carryover: ``\textit{\underline{the one I added earlier}}").
This result suggests future modeling directions that could leverage the unique multimodal context more explicitly.
It can also be shown that the model that uses raw visual embeddings outperforms the model that uses stringfied textual tokens, by better incorporating rich context present in visual information.
%(old result) It can also be shown that the model that uses stringfied textual tokens outperforms the model that uses raw visual embeddings, by better leveraging the causal inference ability of a pre-trained LM.

%% file: tables/tab_model_results.tex
\begin{table}[t]
    \setlength{\tabcolsep}{5pt}
    \begin{center}
        \scalebox{0.75}{%
            \begin{tabular}{
                cccc
            }
            \toprule[\heavyrulewidth]
            \multirow{2}{*}{\textbf{Model}}
            & \textbf{1. API Slot}
            & \textbf{2. Coref}
            & \textbf{3. DST}\\
            % & \textbf{}\\
            \cmidrule(r){2-2}
            \cmidrule(r){3-3}
            \cmidrule(r){4-4}
                & Slot F1$\uparrow$
                & Coref F1$\uparrow$
                & Acc.$\uparrow$\\
            \midrule
            GPT-2 (tokens)
                & \reportval{88.3}{0.3}
                & \reportval{70.4}{0.5}
                & 72.8 \\%\reportval{90.0}{} \\
            GPT-2 (embed)
                & \reportval{\textbf{90.1}}{0.1}
                & \reportval{\textbf{81.5}}{0.6}
                & \textbf{79.6} \\%\reportvl{67.0}{} \\            
            \bottomrule[\heavyrulewidth]
            \end{tabular}
        }
    \end{center}
    %\vspace*{-10pt}
    \vspace*{\captionReduceTop}
     \caption{
        Baseline performances for GPT-2 models w/ multimodal image features (embed) and stringified text (tokens).
        \textbf{(1) API Call Slot Prediction (API Slot)}, via \underline{slot F1}, 
        \textbf{(2) Multimodal Coreference Resolution (Coref)}, via \underline{coref} prediction \underline{F1},
        \textbf{(3) Dialog State Tracking (DST)}, via Joint \underline{Acc}uracy.
        $\uparrow$: higher is better.
        %\textbf{Bold} denotes best performance with statistical significance.
        }
    %\vspace{-10pt}
    \vspace*{\captionReduceBot}
    \vspace*{-4pt}
    \label{tab:results}    
\end{table}

%% file: sections/conclusions.tex
\noindent \textbf{Conclusions}: 
We propose a novel task of building a TOD system for interactively creating storytelling media contents from a personal media collection.
We build a new multimodal dataset ($10k$ dialogs \& $136k$ turns) with rich dialog annotations and story representations.
%, allowing for various modeling approaches.
Our analysis with the SOTA LM-based multimodal dialog model highlights the key challenges such as multimodal coreference resolution and MM-DST.
Lastly, our mobile application demonstrates the feasibility of our \dn dataset and model on popular real-world applications in short and long-form content creation and sharing.

%% file: sections/limitations.tex
The generalizability and the use cases of the \dn{} dataset are bounded by the synthetic nature of the multimodal dialog simulator used for this study.
However, we note that even with the simulated dialog flows, \dn captures several interesting challenges that are not addressed in the previous literature such as the use of media montage representations and device status as the grounding context for multimodal conversations, which opens the door to new research directions.
We will open-source the multimodal dialog simulator used in the study for anyone to further develop any video-editing operations that are not included in \dn, if necessary.

%% file: sections/supplementary.tex
\newpage
\section{Appendix: Demo Interface}
\label{appendix:demo}

\begin{figure*}[b]
%\begin{figure*}[hbtp]
%\begin{minipage}{\textwidth}
  \centering \includegraphics[width=1.98\columnwidth]{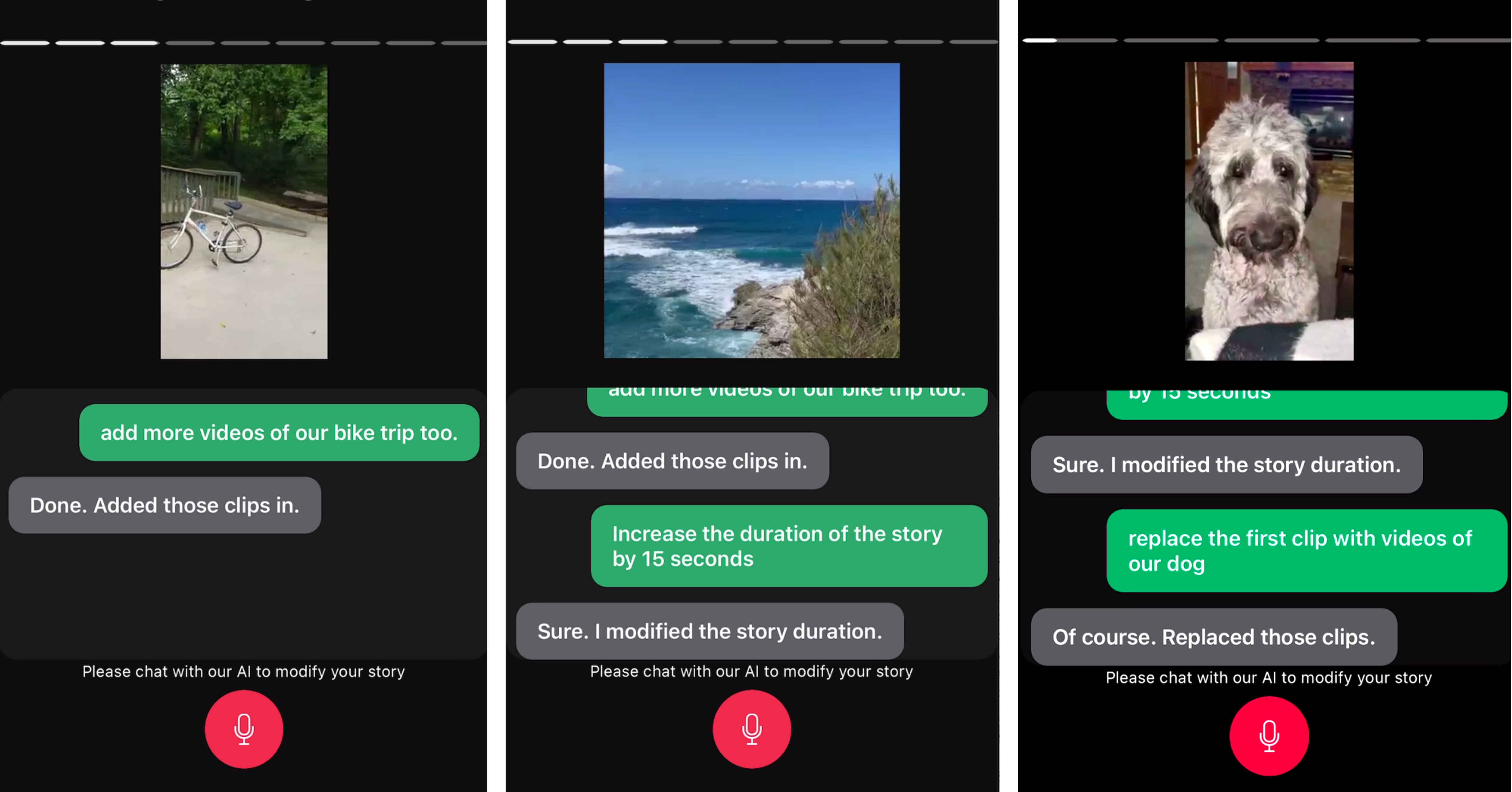}
  %\captionsetup{type=figure}
  %\vspace{-0pt}
    \caption{Screenshots of our \textbf{mobile demo application}. The dialog model is trained with the \dn{} dataset, and served on a Python server. A personal media collection was used as a retrieval target set for demonstration purposes.   }
  \label{fig:demo}
\end{figure*}

To demonstrate the feasibility of the real-world applications of the proposed dataset and models, we built a mobile demo application that runs the model trained with the \dn{} dataset. 
As can be seen in \reffig{fig:demo}, the demo successfully handle unscripted user requests (not drawn from the training data) on a personal video collection as a retrieval target set, showing the promising use cases of our work. 

Note that a computer vision model was used to pre-process and extract key visual concepts for each video in the collection.
Each video was indexed with the extracted concepts and stored in a database in advance for faster inference. 

At inference time, the mobile front-end runs an ASR model to get a transcript of a user's request, which is then routed to the dialog model.
Once the dialog model predicts the API call and parameters, we retrieve the associated video files and execute the requested create or edit operations on the story.

\section{Appendix: Multimodal DST with a Causal Language Model}
\label{appendix:baseline}

Following the recent success of finetuning pretrained LMs on
task-oriented dialog task modeling \cite{simpletod,soloist}, we cast the MM-DST as a causal language inference task.
Specifically, we use the concatenated \{\texttt{<dialog history>, <multimodal context>}\} as the prompting context for the LM (where multimodal context is represented either as visual embeddings or textual tokens), and use the task labels \{\texttt{INTENT [slot = value, ...] <clip: IDs, ...>}\} as the target for causal LM inference.

\begin{figure}
    \centering
    \includegraphics[width=0.99\columnwidth]{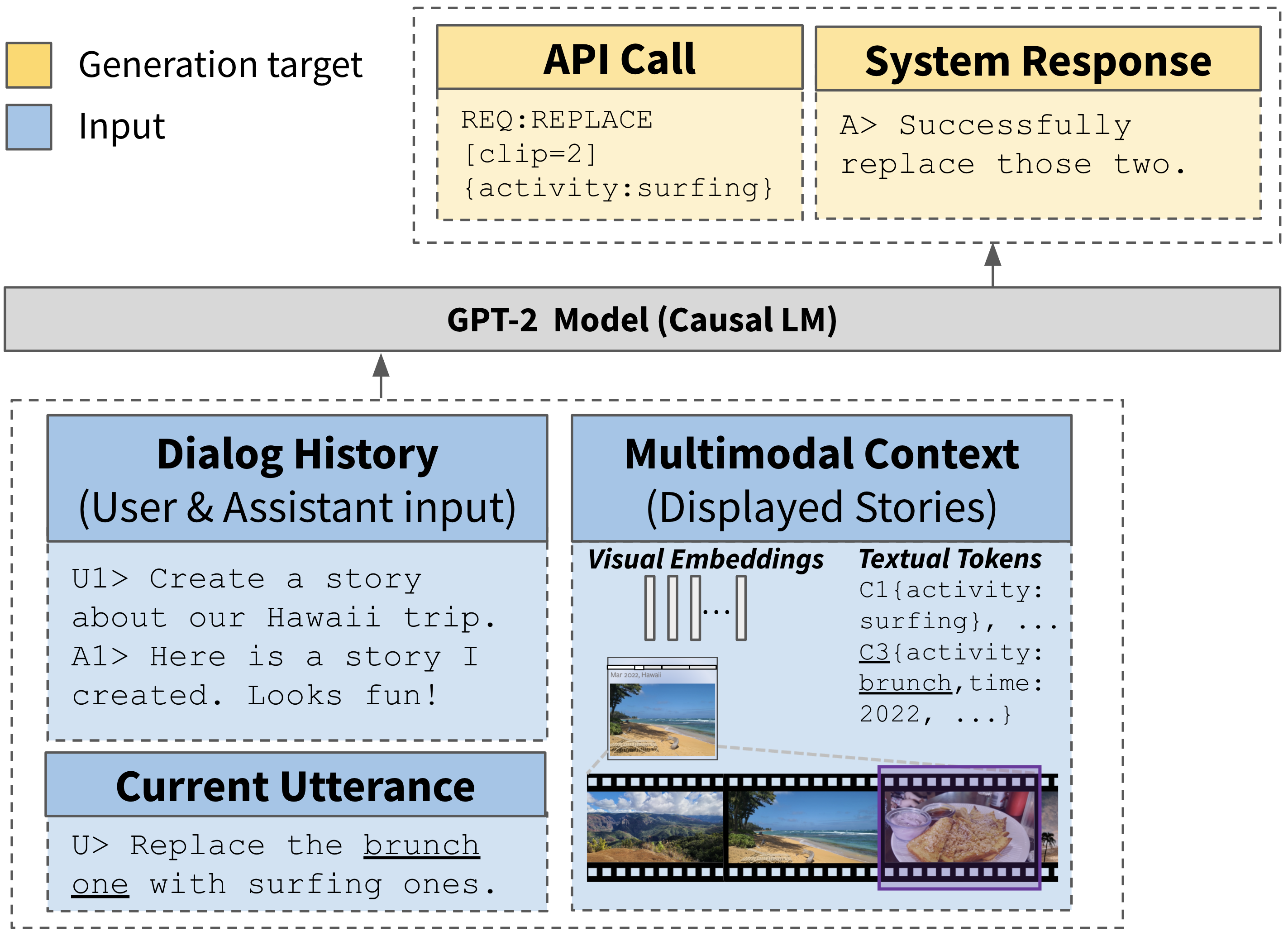}
    \caption{
        Baseline GPT-2 models for \dn. 
        Given the dialog history, multimodal context, and current user utterance, the model
        predicts the API call at the current turn.
        As shown, GPT2 (tokens) uses attribute strings to represent memories, while GPT2 (embed) use visual features.
        % ** Figure made with a slide from: https://docs.google.com/presentation/d/1ZA2bdYg0NtqJOyVVsHoH8tjR9WU7JC1kCimD-7vuLc0/edit#slide=id.g16d790f7210_0_0
    }
    \vspace*{\captionReduceBot} 
    \label{fig:baseline_diagram}
\end{figure}

We use the 12-layer GPT-2 ($117M$) model \cite{radford2019language} and finetune it on the \dn dataset, using early stopping based on token perplexity (<3 GPU hrs).
\reffig{fig:baseline_diagram} illustrates the proposed architecture for the tasks in \refsec{sec:task_formulation}.

\section{Appendix: Dataset Example}
\label{appendix:dataset_example}

\begin{figure*}[b]
%\begin{figure*}[hbtp]
%\begin{minipage}{\textwidth}
  \centering \includegraphics[width=1.98\columnwidth]{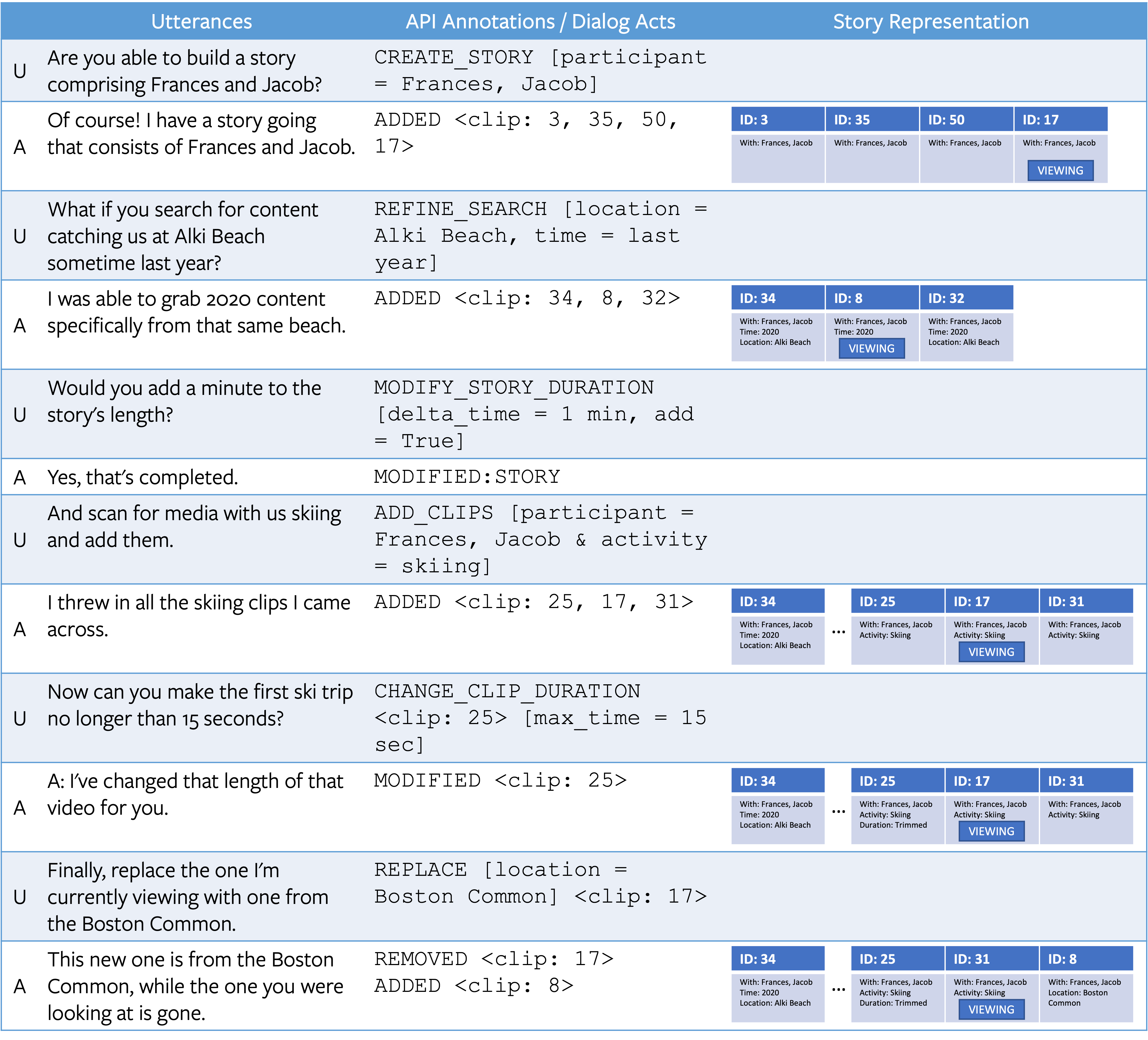}
  %\captionsetup{type=figure}
  %\vspace{-0pt}
    \caption{\textbf{Dataset Example}. Dialog labels include intent, slots, and multimodal coreferences. }
  \label{fig:example}
\end{figure*}

\reffig{fig:example} illustrates an example dialog from the \dn dataset, along with the schematic representation of the stories (with a sequence of clips and their meta data) associated with each turn (U: User, A: Assistant).
API Annotations are formatted as follows: \texttt{INTENT [slot = value, ...] <clip: IDs, ...>}. 

It can be seen that the dataset includes many challenges such as multimodal coreferences and dialog context carryovers. 
We report the detailed breakdown of the benchmark performances (\eg API prediction, Multimodal Coreference Resolution F1) in \refsec{sec:modeling}

More details on the dataset including the key statistics are provided in \refsec{sec:dataset:analysis}.

\section{Appendix: Ethical Considerations}
\label{appendix:ethical_considerations}

The data paraphrase task was contracted through an external vendor that specializes in NLP annotations, where annotators are employed as full-time.
Annotators were provided with clear instructions including a detailed escalation path (``Report Dialog") for an (unlikely) case where the templated utterance may include sensitive topics.

Please note that the figures used in this paper are from authors' personal media collections, and do not include identifiable faces or sensitive topics.

